\documentclass[10pt,twocolumn,letterpaper]{article}

\usepackage[preprint]{cvpr}      %
\usepackage{pifont}

\definecolor{cvprblue}{rgb}{0.21,0.49,0.74}
\usepackage[pagebackref,breaklinks,colorlinks,allcolors=cvprblue]{hyperref}
\usepackage{multirow}
\usepackage[table]{xcolor}
\usepackage{tcolorbox}
\usepackage{wrapfig}

\newcounter{alphasect}
\def\alphainsection{0}

\let\oldsection=\section
\def\section{%
  \ifnum\alphainsection=1%
    \addtocounter{alphasect}{1}
  \fi%
\oldsection}%

\renewcommand\thesection{%
  \ifnum\alphainsection=1%
    \Alph{alphasect}%
  \else
    \arabic{section}%
  \fi%
}%

\def\paperID{xx} %
\def\confName{CVPR}
\def\confYear{2026}

\title{\textsc{RetouchIQ}: MLLM Agents for Instruction-Based Image Retouching with Generalist Reward}

\author{
\textbf{Qiucheng Wu}${^{1,2}}\thanks{This work was completed during Qiucheng’s internship at Adobe.}$, \textbf{Jing Shi}$^{1}$,
\textbf{Simon Jenni}${^1}$, \textbf{Kushal Kafle}${^1}$, \textbf{Tianyu Wang}${^1}$, \textbf{Shiyu Chang}${^2}$, \textbf{Handong Zhao}${^1}$\thanks{Project Lead.}\\
$^1$Adobe Research, $^2$UC, Santa Barbara}

\begin{document}

\let\oldtwocolumn\twocolumn
\renewcommand\twocolumn[1][]{%
    \oldtwocolumn[{#1}{
    \begin{center}

\vspace{-1em}
\includegraphics[width=0.97\textwidth]{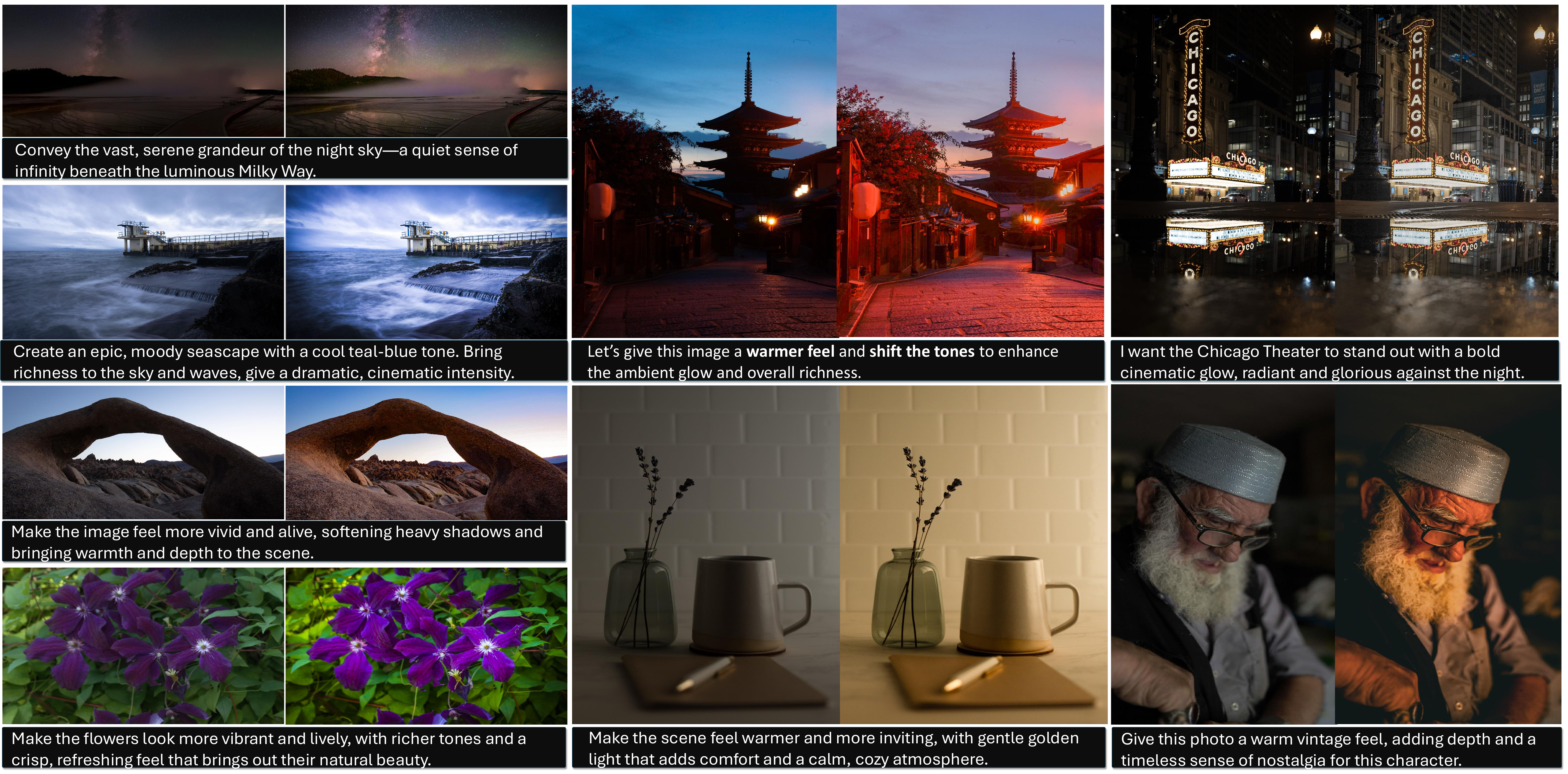}
\vspace{-0.5em}
    \captionof{figure}{We present \textsc{RetouchIQ}, an MLLM agent that performs customized image retouching. Given an user instruction and an image input, \textsc{RetouchIQ} produces high-quality results across various quality enhancement and style transformation scenarios.}
    \label{fig:intro}
        \end{center}
    }]
}

\maketitle
\begin{abstract}
Recent advances in multimodal large language models (MLLMs) have shown great potential for extending vision-language reasoning to professional tool-based image editing, enabling intuitive and creative editing. 
A promising direction is to use reinforcement learning (RL) to enable MLLMs to reason about and execute optimal tool-use plans within professional image-editing software. However, training remains challenging due to the lack of reliable, verifiable reward signals that can reflect the inherently subjective nature of creative editing.
In this work, we introduce \textsc{RetouchIQ}, a framework that performs instruction-based executable image editing through MLLM agents guided by a generalist reward model.
\textsc{RetouchIQ} interprets user-specified editing intentions and \textcolor{black}{generates corresponding, executable image adjustments}, bridging high-level aesthetic goals with precise parameter control.
\textcolor{black}{To move beyond conventional, rule-based rewards that compute similarity against a fixed reference image using handcrafted metrics}, 
we propose a generalist reward model—an RL fine-tuned MLLM that evaluates edited results through a set of generated metrics on a case-by-case basis. 
Then, the reward model provides scalar feedback through multimodal reasoning, enabling reinforcement learning with high-quality, instruction-consistent gradients. 
We curate an extended dataset with 190k instruction-reasoning pairs and establish a new benchmark for instruction-based image editing. Experiments show that \textsc{RetouchIQ} substantially improves both semantic consistency and perceptual quality over previous MLLM-based and diffusion-based editing systems.
Our findings demonstrate the potential of generalist reward-driven MLLM agents as flexible, explainable, and executable assistants for professional image editing.
\end{abstract}
    
\section{Introduction}
\label{sec:intro}
The recent progress of multimodal large language models (MLLMs) has unlocked a vast landscape of applications that bridge visual understanding and linguistic reasoning. Beyond perception and captioning, MLLMs are now capable of acting as general-purpose agents that can invoke external tools, combining the broad reasoning capabilities of language models with the precision and domain expertise of specialized systems. Such MLLM agents have demonstrated their potential in a wide range of domains, from retrieval~\cite{wang2025vrag,openai-deepresearch} and image generation~\cite{wang2024genartist,yang2024mastering,huang2024genmac} to autonomous workflows~\cite{lu2025ui,luo2025gui,openai-agent,gemini-agent}, showcasing an emerging paradigm of reasoning-driven tool use.

Among these applications, one particularly creative and artistically meaningful direction is to empower MLLM agents to interact with professional image editing software, \textcolor{black}{such as Adobe Lightroom, PicsArt, and Claid AI studio}~\cite{adobe-lightroom,picsart,claid}.
However, mastering such software demands deep technical and creative expertise, posing a barrier for both casual users and professional photographers who may excel at capturing images but lack the specialized knowledge for effective retouching. Existing solutions~\cite{dutt2025monetgpt,lin2025jarvisart,chen2024restoreagent} adopt various post-training strategies to equip MLLMs with tool-use capabilities, including reinforcement learning (RL) aimed at reproducing human-edited results. However, due to the inherently subjective nature of creative photo retouching, multiple edits can be equally good, and different users may produce distinct yet valid outcomes for the same request. Consequently, existing rule-based pixel-level rewards (See Figure~\ref{fig:reward_problem}) become unreliable when model training is anchored to a single reference edit.

To address this challenge, we introduce \textsc{RetouchIQ}, an MLLM-based agent that performs instruction-based executable image editing, converting user requests into interpretable reasoning and actionable steps. \textcolor{black}{\textsc{RetouchIQ} leverages editing tools from professional image processing software} to execute fine-grained parameter adjustments that align with natural language instructions (e.g., ``create an epic, moody seascape, give a dramatic, cinematic intensity'' as shown in Figure~\ref{fig:intro}). 
Crucially, we introduce a generalist reward model (GRM), which is an RL fine-tuned MLLM that evaluates edited results with a series of generated metrics on a case-by-case basis. 
To construct pairwise (strong/weak) data for RL training of the GRM, a natural approach is to generate weak edits via inverse-edit perturbations—using predefined tools and parameter offsets~\cite{conde2025pixtalk}.
However, we find that this seemingly reasonable strategy yields sub-optimal performance due to a distribution shift between these synthetic weak edits and the real weak edits produced by the policy editing model. We further introduce a new training scheme (Policy-Guided Reward Training, PGRT) designed to align the data distributions of the policy editing model and the reward critic model, thereby yielding stronger and more stable performance.

Experiments show that \textsc{RetouchIQ} exhibits strong capabilities in producing aesthetically consistent, instruction-aligned, and publication-quality image edits. Compared with existing agent and diffusion-based editing systems, our framework offers better semantic consistency and perceptual quality, while providing explainable reasoning throughout the editing process. Our contributions are threefold: 

\begin{itemize}
\item \textcolor{black}{We introduce \textsc{RetouchIQ}, the first framework to address the inherently subjective nature of image retouching by introducing a \textit{generalist reward model} that provides flexible, context-aware supervision beyond conventional verifiable rewards.}
\item \textcolor{black}{We propose PGRT, a RL paradigm that leverages signals from the policy editing model itself to refine reward estimation and better capture nuanced perceptual differences between before- and after-edit images.}
\item \textsc{RetouchIQ} demonstrates strong performance in both semantic alignment and perceptual quality, surpassing state-of-the-art general purpose MLLMs, MLLM agents and diffusion-based systems, and producing edits that closely approach professional-level retouching.
\end{itemize}

\section{Related Work}
\subsection{Image Retouching}
Image retouching aims to enhance visual appeal by adjusting tone, color, and illumination while preserving realism and semantic consistency.
Early works focused on learning to predict image editing operations to achieve aesthetically pleasing results~\cite{kosugi2020unpaired,ke2022harmonizer,hu2018exposure,ouyang2023rsfnet,shi2021learning}.
While these methods provide step-wise and explainable editing procedures, these models typically have little or no capability to capture nuanced, user-specific goals, which prevents them from handling diverse textual instructions.
More recently, diffusion-based models~\cite{ho2020denoising,song2020score} have been explored for image enhancement and editing according to user prompts~\cite{duan2025diffretouch,wu2023uncovering,liu2025mofrr,hertz2022prompt}.
Although these models can interpret natural-language instructions, they often unintentionally alter the original image content and environment due to the stochastic nature of the diffusion process.
To address these limitations, recent research has begun exploring multimodal LLM agents that interact with external editing tools to achieve desired retouching effects~\cite{dutt2025monetgpt,lin2025jarvisart,chen2024restoreagent}.
These studies demonstrate that MLLMs possess potential to understand user intentions and reason about proper editing steps, even in a zero-shot setting~\cite{ye2025retouchllm,he2024training}.

\begin{figure*}
    \centering
    \includegraphics[width=0.94\textwidth]{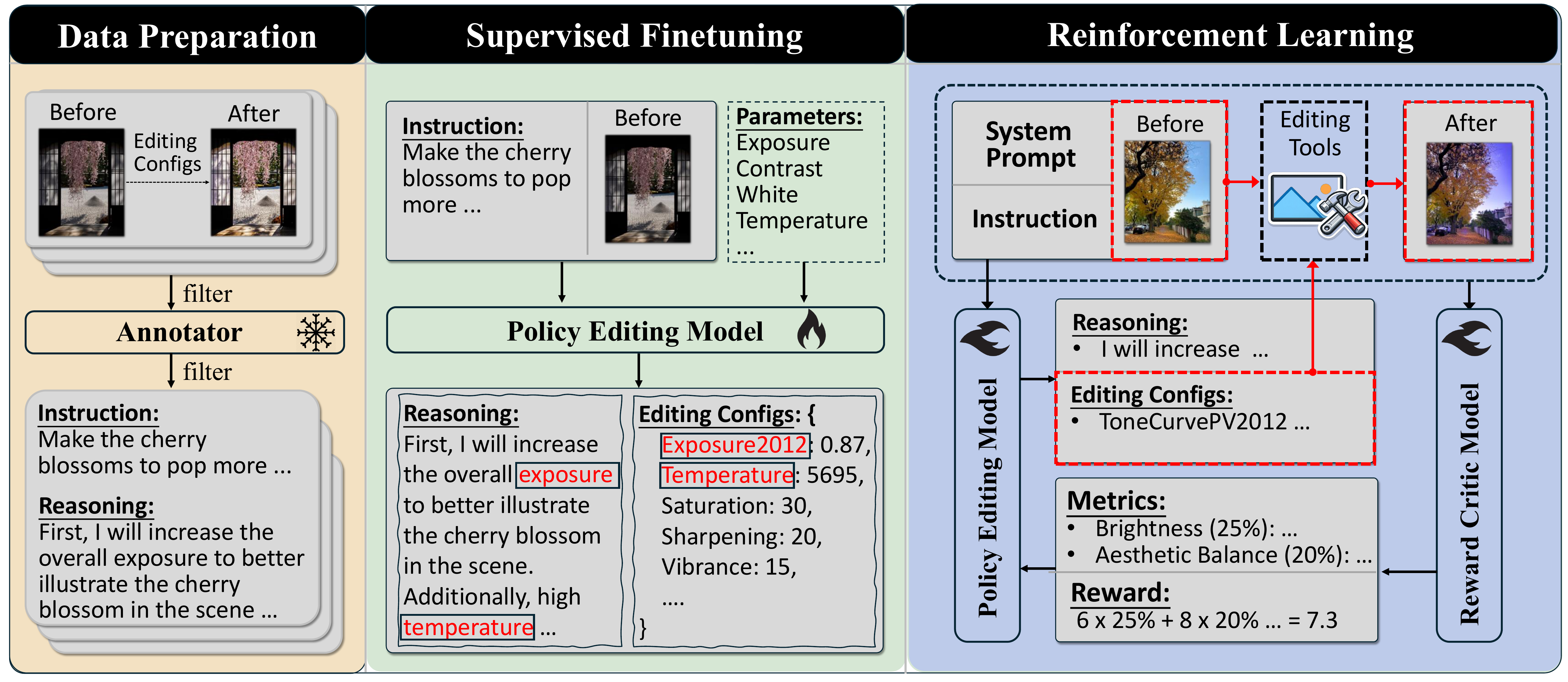}
    \vspace*{-0.1in}
    \caption{\textcolor{black}{Overview of \textsc{RetouchIQ}. \textbf{Left:} We annotate the user instruction and reasoning for training data. Generated data are filtered to ensure quality. \textbf{Middle:} The supervised fine-tuning stage. Based on the user instruction (\emph{e.g.,} pop more), the policy model needs to reason the correct parameters (\emph{e.g.,} \texttt{exposure}) and change them accordingly. \textbf{Right:} The reinforcement learning stage. We leverage a generalist reward model to propose metrics and provide scalar reward guidance for policy model. Details of the reward model is introduced in Sec~\ref{sec:GRM}.}}
    \vspace*{-0.15in}
    \label{fig:method}
\end{figure*}

\subsection{Generalist Reward in Reinforcement Learning}
Precise reward signals are central to fine-tuning LLMs for reasoning capabilities.
Previous approaches mainly rely on verifiable rewards~\cite{guo2025deepseek,hong2025glm,wang2025vl}, which provide rule-based scores determined by the correctness of the final answers.
While such methods have proven effective in objective domains such as mathematics and programming~\cite{shen2025skywork,qiao2025we,fan2025sophiavl}, recent studies show that they struggle in more complex and subjective tasks~\cite{liu2025inference}, such as the agentic scenario.
One way to overcome this is through a \textit{generalist reward model}~\cite{zhou2025generative,liu2025inference}, where an LLM reward model is trained to score the answers.
In this work, we extend this concept by training an MLLM-based reward model that assesses output images through a series of dynamically generated evaluation metrics.

\section{\textsc{RetouchIQ}}
\label{sec:method}
In this section, we introduce \textsc{RetouchIQ}, an MLLM agent that performs instruction-based image retouching leveraging professional editing tools. We first introduce the overall setup in Sec.~\ref{sec:method_overview}. Then, we introduce the training pipeline, including data preprocessing (Sec.~\ref{sec:method-data}), supervised fine-tuning (Sec.~\ref{sec:method-sft}), and reinforcement learning (Sec.~\ref{sec:method-rl}). 

\subsection{Overview}\label{sec:method_overview}

The \textsc{RetouchIQ} pipeline is illustrated in Figure~\ref{fig:method}. 
\textsc{RetouchIQ} includes a policy model that translates a user’s natural-language instruction into two outputs:
\begin{itemize}
    \item A \textbf{reasoning trace}, describing the semantic interpretation and aesthetic intent of the request, and
    \item A sequence of \textbf{parameterized editing tool-use operations} that can be executed (e.g., \{\texttt{exposure=+0.9}; \texttt{contrast=-30}\}).
\end{itemize}
To achieve this, we adopt a two-stage training strategy.
In the first stage, supervised fine-tuning (SFT), the model learns to simulate 
\textit{gold} reasoning processes and corresponding editing operations from real user edits.
In the second stage, reinforcement learning (RL), the model self-explores diverse reasoning paths and editing plans, gradually discovering better editing strategies under different input conditions.
However, collecting a suitable training dataset that includes images, user editing intentions, and detailed editing steps required for SFT and RL remains challenging. To address this, we design a data collection pipeline to curate and preprocess high-quality data for subsequent training.

\subsection{Data Preparation}\label{sec:method-data}
In \textsc{RetouchIQ}, an ideal training datum should contain the following elements: a before-and-after image pair $(I_{0}, I)$, a user editing goal $g$ expressed in natural language, a reasoning process $q$ describing how to plan the optimal editing strategy, and an editing step sequence $e$ consisting of parameter configurations.

We present the data preparation pipeline in the left panel of Figure~\ref{fig:method}. \textcolor{black}{We start by collecting initial data from user editing trajectories, where the before-and-after image pairs $(I_{0}, I)$ along with the corresponding editing sequences $e$ are recorded.} Note that, unlike existing work~\cite{lin2025jarvisart} that synthesizes post-edit images by applying predefined editing presets, our before–after image pairs are collected from real human users \textcolor{black}{(all of whom provided the required permissions for data use)}, making them naturally closer to real-world editing scenarios. However, two essential elements are missing from the source data: \ding{182} the specific editing goal $g$ that the user had when modifying the image, and \ding{183} the reasoning process $q$ that guided the user’s use of corresponding image editing tools. 

As such, we seek to annotate $g$ and $q$ by leveraging an MLLM-based annotator. Specifically, the annotator takes as input the triplet $(I_{0}, I, e)$ and first infers the user's editing intention $g$ based on the differences between the images and the applied editing steps. It is then prompted to generate a reasoning process $q$, simulating how an MLLM agent would reason and produce the final editing sequence $e$ accordingly.
To ensure the quality of the generated instructions and reasoning processes, we apply a systematic filtering procedure to remove image pairs with unclear editing intentions or inconsistencies between the generated intentions and reasoning. The MLLM annotator remains fixed throughout the entire process.

\subsection{Supervised Fine-Tuning Stage}\label{sec:method-sft}
With the prepared training data, we now perform supervised fine-tuning of the MLLM policy model in \textsc{RetouchIQ}.
We initialize the policy model from a pretrained MLLM backbone and fine-tune it using the instruction–reasoning–edit corpus.
Given an input image $I_0$ and user instruction $g$, the model is trained to generate both a textual reasoning trace $q$ and structured editing steps $e$.
The supervised fine-tuning objective is an autoregressive loss that minimizes the negative log-likelihood over the target token sequence:
\begin{equation}
\mathcal{L}_{\text{SFT}} = - \sum_{t} \log p_{\theta}(y_t \,|\, y_{<t},\, I_0, g),
\label{eq:sft}
\end{equation}
where $\theta$ denotes the model parameters, $y_t$ is the $t$-th output token, and $y_{<t}$ represents all previously generated tokens.
The output sequence contains both the natural-language reasoning $q$ and the editing operations $e$ in a structured format.
This stage enables the model to learn semantically grounded mappings between linguistic aesthetic intent and executable editing parameters, providing a strong initialization for downstream reinforcement learning.

\subsection{Reinforcement Learning Stage}\label{sec:method-rl}
While supervised fine-tuning aligns the policy editing model with demonstration data, it cannot guarantee optimal aesthetic alignment with user intent across diverse images and instructions.
To further improve the model’s ability to generalize and adapt, we refine the policy through RL, allowing it to explore various reasoning paths and editing strategies beyond the demonstration data.

\begin{figure}[t]
    \centering
    \includegraphics[width=\columnwidth]{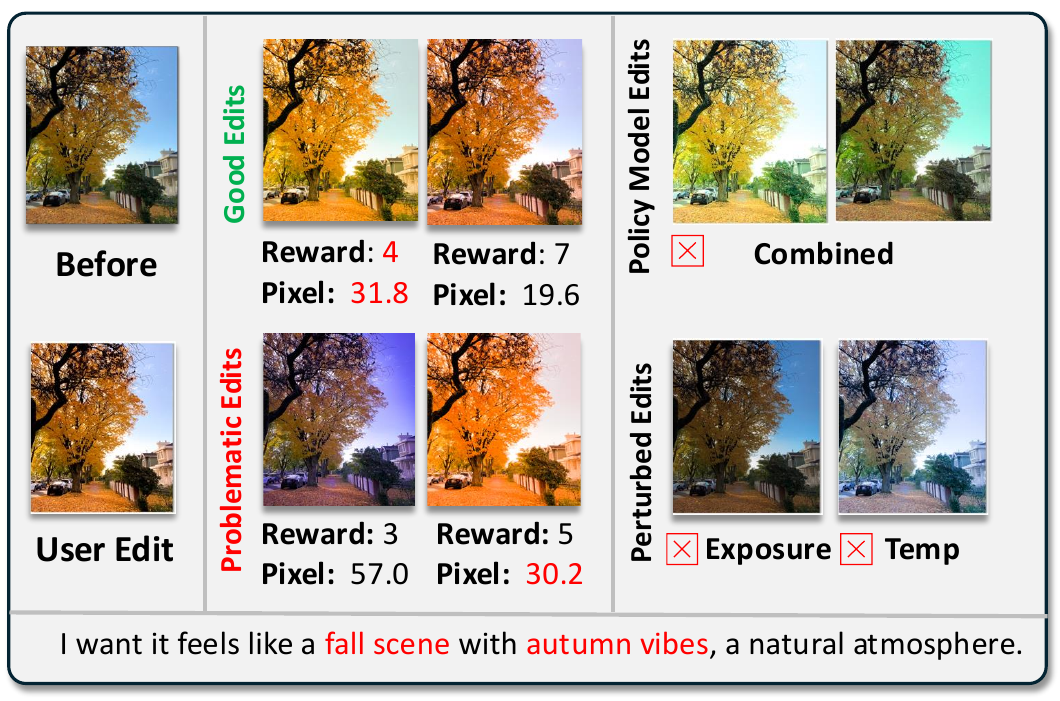}
    \vspace*{-0.2in}
    \caption{Problematic rewards in image retouching tasks. Given a before–after image pair (\textbf{left}), \ding{182} \textbf{verifiable rewards} (\textbf{middle}) rely on metrics between the edited image and ground truth, such as pixel differences. However, since multiple valid edits can satisfy user intent, these rewards become imprecise. \ding{183} The \textbf{reward model}'s precision strongly depends on its \textbf{training data distribution} (\textbf{right}). When trained to distinguish good user edits from randomly perturbed images, it may later struggle to assess results from the policy model that produces combined, complex edits.} 
    \vspace*{-0.125in}
    \label{fig:reward_problem}
\end{figure}

\begin{figure*}
    \centering
    \includegraphics[width=0.95\textwidth]{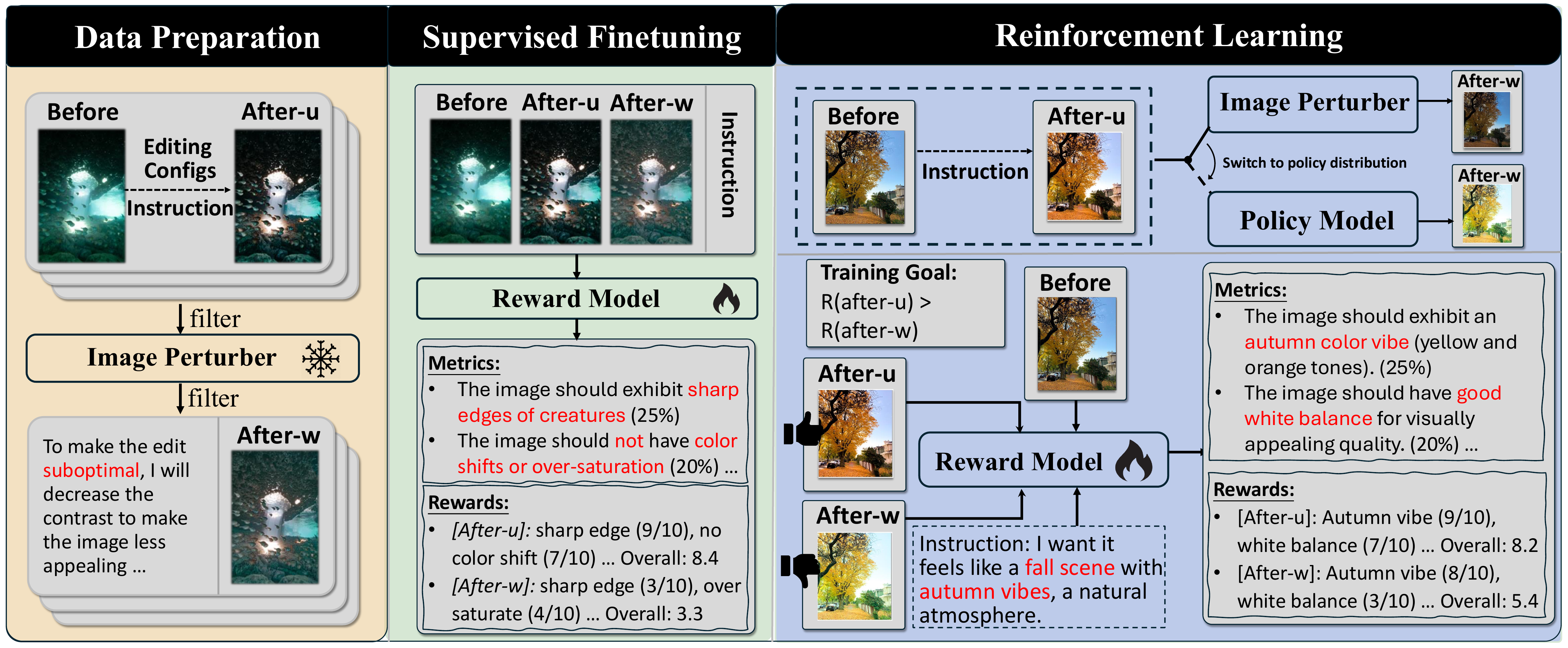}
    \vspace*{-0.05in}
    \caption{
    Overview of generalist reward model. \textbf{Left}: Given a before-edited image and a user-edited after image \texttt{After-u} (``u'' for ``user''), we perturb the editing process to obtain a suboptimal image, \texttt{After-w} (``w'' for ``weak''). \textbf{Middle}: In the SFT stage,the reward model first learns to generate metrics and then produces scalar rewards for the two input images. \textbf{Right}: In the RL stage, we introduce policy-guided reward training (PGRT), where the suboptimal image \texttt{After-w} is provided by the policy model rather than generated through perturbation. The goal of the RL stage is to assign a higher scalar reward to the user-edited (strong) image \texttt{After-u}.}
    \vspace*{-0.15in}
    \label{fig:grm}
\end{figure*}

A conventional practice in RL is to develop a verifiable reward, which gives score based on pre-defined rules such as similarities to a ground truth reference image~\cite{lin2025jarvisart}.
However, this strategy faces a fundamental challenge in image editing, where \textit{the inherently subjective nature of aesthetics makes it impossible to define a single verifiable ground truth}. As illustrated in Fig.~\ref{fig:reward_problem}, a given instruction (e.g., ``make it feel like a fall scene with autumn vibes'') can correspond to multiple acceptable results, each expressing the intent differently through tone, warmth, or color balance. In this setting, conventional reward definitions based on pixel-wise or reference-based comparisons fail to capture semantic quality or artistic preference. Such metrics are suitable for deterministic reconstruction tasks but are inadequate for open-ended, instruction-driven retouching, where subjective aesthetic judgments play a central role.

We therefore further refine the policy via RL using the \textit{reward model} $r_{\phi}$, which evaluates how well an edited result satisfies the given instruction.
For a policy $\pi_{\theta}(q,s \,|\, I_0, g)$ that predicts editing actions given the image and instruction, the objective is to maximize the expected reward:
\begin{equation}
\mathcal{J}(\theta) = 
\mathbb{E}_{q,s \sim \pi_{\theta}}
\left[
r_{\phi}\!\left(g,\, I_0, \text{Execute}(I_0, e)\right) + r_{\text{format}}(q,s)
\right],
\label{eq:rl_obj}
\end{equation}
where $\text{Execute}(\cdot)$ denotes applying the predicted edits to obtain the resulting image, and $r_{\text{format}}$ is a format reward~\cite{guo2025deepseek}. During training, the policy samples candidate edits, receives feedback, and updates its parameters via gradient ascent on $\mathcal{J}(\theta)$, progressively improving instruction alignment and aesthetic quality.

As such, the success of this RL stage heavily depends on the quality of the reward model. Ideally, the reward model should provide a scalar signal that accurately reflects the quality of the generated edit. However, the criteria for a successful edit vary across different input images and user instructions. To flexibly adapt to diverse scenarios and deliver precise reward guidance, we propose a \textit{generalist reward model} that first generates a set of metrics describing the key aspects of a successful edit and then scores the output image accordingly. The following section provides details of the generalist reward model.

\section{Generalist Reward Model}
\label{sec:GRM}
In this section, we present the Generalist Reward Model (GRM) used in \textsc{RetouchIQ}, describing the preparation of its training data and the procedures for model optimization.

\subsection{Overview}
The overall GRM preparation pipeline is shown in Figuer~\ref{fig:grm}. The proposed MLLM-based reward model evaluates image retouching quality by taking a before–after image pair $(I_0, \text{Execute}(I_0, e))$ and the corresponding editing instruction $g$, the reward model produces two outputs sequentially:
\begin{itemize}
\item A self-generated set of \textbf{metrics} that describe the key characteristics a successful edit should exhibit, and
\item A scalar \textbf{reward} value reflecting the overall quality of the retouching according to these metrics.
\end{itemize}
Similar to the policy model, we first prepare dedicated training data for the GRM.
We then adopt SFT and RL to ensure that the GRM provides accurate and consistent guidance across diverse retouching scenarios.

\subsection{Data Preparation}\label{sec:perturb-data}
When preparing the training data, the key challenge lies in constructing accurate and reliable \textit{gold} signals.
Our primary objective is to ensure the correct ordering in supervision: a better retouching result should receive a higher reward than a suboptimal one under the same instruction.

To achieve this, we construct a paired comparison dataset by augmenting the training corpus with \emph{strong} and \emph{weak} edits for each original image, as shown in the left panel of Figure~\ref{fig:grm}.
Starting from the initial data pool containing before-after pairs $(I_0, I)$, we treat the reference after image $I$ as a strong edit corresponding to the annotated instruction $g$.
To obtain a weak edit, we intentionally \textit{perturb} the editing configuration $s$ to produce an alternative version that omits or misadjusts several key parameters during the editing process, resulting in a suboptimal outcome $I_w$.
This perturbation is guided by a frozen MLLM to maintain edits that remain plausible yet perceptually inferior.
We further apply filtering mechanisms to ensure that each generated $I_w$ is indeed weaker than $I$, providing reliable ranking consistency for reward learning. 

Finally, we annotate each $(I, I_w)$ pair using an MLLM.
Since the ground-truth ordering between the strong and weak edits is known, we retain only annotations that correctly assign a higher score to $I$, obtaining annotated metrics $m$ and scores for the strong and weak response $(s, s_w)$ correspondingly.

\subsection{Training}
\paragraph{Supervised Fine-Tuning Stage.}
We first perform supervised fine-tuning so the GRM learns to produce metrics and corresponding scores consistent with annotations.
The training examples contain input image $I_0$, editing instruction $g$, two after edit image $I$ and $I_w$, and annotated metric $m$ and scores $(m, s, s_w)$. The training objective is an autoregressive loss that fits the annotation:
\begin{equation}
\mathcal{L}_{\text{SFT}}^{\text{reward}} = - \sum_{t} \log p_{\phi}(y_t \,|\, y_{<t},\, I_0, I, I_w, g).
\label{eq:reward_sft}
\end{equation}
This stage trains the model to explain aesthetic metrics in natural language and to produce scalar scores accordingly. 

\paragraph{Policy-Guided Reward Training (PGRT) in RL.}
The previous SFT stage relies on perturbed sub-optimal images, which introduces intrinsic bias toward a limited set of edit categories. As illustrated in Figure~\ref{fig:reward_problem}, we observe that many perturbations involve single adjustments to \texttt{exposure} and \texttt{temperature}, while the results generated by policy model often involve combined and complex edits. This reveals a training distribution shift, which may cause inferior performance of the reward model when predicting real outputs from policy models.

Our goal is to align the reward model with the policy distribution during training. To this end, we propose \textit{policy-guided reward training} (PGRT), which trains the reward model on data consistent with the policy’s editing distribution. As shown in Figure~\ref{fig:grm} (right), the RL stage replaces perturbed images with policy-generated results, shifting $I_w$ from the perturbed to the policy-generated distribution. Given paired examples $(I_0, I, I_w)$, the reward function is defined as
\begin{equation}
\mathcal{J}(\phi) = \mathbb{E}_{m, r, r_w \sim \pi{\phi}}
\left[
\mathbb{I}[r > r_w] + r_{\text{format}}(m, r, r_w)
\right],
\label{eq:reward_rl}
\end{equation}
where $\mathbb{I}[\cdot]$ is a 0-1 indicator, and $r_{\text{format}}$ is a format reward ensuring that both metrics and scores are properly generated.
Essentially, this reward function penalizes the model when it incorrectly assigns a higher score to a weaker edit.

PGRT relies on the outputs of the policy model. Therefore, when training \textsc{RetouchIQ}, we adopt an alternating training strategy in which the policy model and the reward model are optimized in turn to progressively enhance each other’s performance. It is worth noting that when training the reward model for the first time, the RL stage uses only perturbed edits, since the policy model is not yet trained.

\begin{table*}
    \centering
    \resizebox{\linewidth}{!}{%
\begin{tabular}{lccccccccccccccc}
\toprule
\multicolumn{1}{c}{\multirow{3}{*}{\textbf{}}} & \multicolumn{5}{c}{\textbf{Quality Improving}} & \multicolumn{5}{c}{\textbf{Style Changing}} & \multicolumn{5}{c}{\textbf{Local Retouching}}\\

\cmidrule(lr){2-6}
\cmidrule(lr){7-11}
\cmidrule(lr){12-16}

  & L1 $\downarrow$ & L2 $\downarrow$ & SC $\uparrow$ &  PQ $\uparrow$ & O $\uparrow$ & L1 $\downarrow$ & L2 $\downarrow$ & SC $\uparrow$ & PQ $\uparrow$ & O $\uparrow$ & L1 $\downarrow$ & L2 $\downarrow$ & SC $\uparrow$ & PQ $\uparrow$ & O $\uparrow$ \\

\midrule
\textsc{Flux-Pro}~\citep{flux-1d1} &  64.97 & 98.14 & 6.12 & 6.09 & 6.10 & 68.20 & 70.01 &  6.48 &  6.36 & 6.42 & 67.93 & 82.86 &  6.19 & 6.26 & 6.22 \\
\textsc{GPT-5}~\citep{gpt5} & 36.21 & 47.53 & 6.74 & 6.51 & 6.62 & 39.17 & 50.36 & 6.89 & 6.76 & 6.82 &  37.20 & 47.56 & 6.07 & \underline{6.89} & 6.47 \\
\textsc{Gemini-2.5}~\citep{comanici2025gemini} & 39.58 & 51.60 & 6.56 & 6.72 & 6.64 & 41.54 & 53.01 & 5.82 & 6.28 & 6.05 & 36.11 & 46.72 &  6.01 & 6.63 & 6.31
  \\
\textsc{MonetGPT}~\citep{dutt2025monetgpt} & 32.05 & \textbf{44.98} & 6.52 & \underline{7.05} & 6.78 &  41.03 & 52.75 & 5.87 & 6.03 & 5.95 & 30.08 & 44.98 & 6.27 & 6.70 & 6.48 \\ 
\textsc{JarvisArt}~\citep{lin2025jarvisart} & \underline{32.17}  & 46.23  & \underline{7.22}  & 6.59 & \underline{6.90} & \textbf{33.90} & \underline{48.23} & \textbf{7.39} & 6.88 & \underline{7.13} & 38.19 & 49.75 & \textbf{6.45} & 6.39 & 6.42 \\ 
\rowcolor{gray!20}
\textsc{RetouchIQ}-SFT  &   35.03 & 46.63 & 6.71 & 6.67 & 6.69 & 38.80 & 51.55 & 7.04 & \underline{6.96} & 7.00 & \textbf{26.41} & \textbf{42.10} & 5.98 & 6.84 & 6.40 \\
\rowcolor{gray!20}
\textsc{RetouchIQ}-Rule  & 32.65 & 46.47 & 7.14 & 6.61 & 6.87 & 35.74 & 49.01 & 7.17 & 6.91 & 7.04 & 28.22 & 43.89 & 6.20 & 6.83 & \underline{6.51} \\
\rowcolor{gray!20}
\textsc{RetouchIQ}-GRM  &  \textbf{31.41} & \underline{44.99} & \textbf{7.57} & \textbf{7.48} & \textbf{7.51} &  \underline{34.08} & \textbf{47.71} & \underline{7.29} & \textbf{7.34} & \textbf{7.31}  & \underline{27.03} & \underline{42.73} & \underline{6.39} & \textbf{6.92} & \textbf{6.65} \\

\bottomrule

\end{tabular}
}
\vspace*{-0.5em}
    \caption{Quantitative comparisons on the \textsc{RetouchEval} benchmark. We consider three categories of baselines: general-purpose MLLMs (GPT-5, Gemini-2.5); MLLM agents (MonetGPT, JarvisArt); and diffusion-based methods (Flux-Pro). Meanwhile, we report the performance of \textsc{RetouchIQ} under both the SFT (RetouchIQ-SFT) and RL (RetouchIQ-RL) stages. We also include results from a variant trained with rule-based reward (RetouchIQ-Rule). The best results are shown in \textbf{bold}, and the second-best results are \underline{underlined}.} 
\label{tab:objective_eval}
\vspace*{-0.5em}
\end{table*}

\section{Experiments}
In this section, we evaluate \textsc{RetouchIQ} under various retouching settings and compare it with state-of-the-art methods.
We focus on the following questions:
\begin{itemize}
    \item Can \textsc{RetouchIQ} consistently provides high-quality image retouching results?
    \item How does the generalist reward model overcome the challenges in conventional verifiable reward?
    \item How does the policy-guided reward training improves the reward model and policy model in \textsc{RetouchIQ}?
\end{itemize}

\subsection{Experiment Settings}
\paragraph{Datasets.}
We curate large-scale, high-quality datasets for training both the policy model and the reward model.
For the policy model, the final training set contains 190K image–instruction pairs covering a wide range of \textit{global} and \textit{local} retouching tasks.
Each input image is associated with three instruction variants that differ in length and complexity, providing diversity during training.
For the reward model, we construct a dataset of 10K perturbed samples, where each datum consists of an original image, its ground-truth editing, and a corresponding sub-optimal version.
During alternating training, an additional 5K samples are generated to further fine-tune the reward model under the RL objective.

\noindent\textbf{Implementation.}
Both the policy and reward models are built upon \texttt{Qwen2.5-VL-7B}~\cite{bai2025qwen2}.
For the MLLM annotator and perturber, we employ \texttt{GLM-4.5V}~\cite{hong2025glm}.
During training, we first fine-tune the reward model using the perturbed dataset.
We then perform alternating fine-tuning of the policy and reward models, as described in Sec.~\ref{sec:method} and ~\ref{sec:GRM}.
\textcolor{black}{For the editing platform, following prior work~\cite{lin2025jarvisart}, we leverage Adobe Lightroom~\cite{adobe-lightroom} for its professional-grade image retouching capabilities.}

\noindent\textbf{Benchmarks.} To the best of our knowledge, there is no existing public benchmark for evaluating image retouching quality using professional editing tools.
Therefore, we introduce \textsc{RetouchEval}, a dedicated benchmark designed to assess retouching quality.
\textsc{RetouchEval} consists of 300 instruction–image pairs curated from real user editing histories.
The dataset is divided into three splits: \textit{quality enhancement}, which focuses on enhancing the overall visual quality of the input image; \textit{style transformation}, which aims to alter the artistic or tonal style of the image; and \textit{local retouching}, which targets localized adjustments rather than global edits.
\textcolor{black}{In addition, we also evaluate on MIT-Adobe5K~\cite{fivek}, a public benchmark for general image enhancement.}
It is worth noting that, unlike our setting, MIT-Adobe5K does not include explicit user intentions; as its primary objective is to improve perceptual quality rather than to achieve a customized editing style. Following previous work~\cite{dutt2025monetgpt}, we conduct evaluations on 400 randomly selected test images.

\noindent\textbf{Baselines.} We compare \textsc{RetouchIQ} with three categories of state-of-the-art baselines:
\ding{182}~\underline{\textbf{General-purpose MLLMs}}: We carefully design editing instructions and prompt leading MLLMs, including \texttt{Gemini-2.5}~\cite{comanici2025gemini} and \texttt{GPT-5}~\cite{gpt5}, to evaluate their zero-shot ability to interpret user instructions and generate valid editing plans.
\ding{183}~\underline{\textbf{MLLM Agents}}: \textcolor{black}{\texttt{JarvisArt}~\cite{lin2025jarvisart} serves as a representative state-of-the-art agent that leverages tools for image editing.}
We also include \texttt{MonetGPT}~\cite{dutt2025monetgpt}, a strong baseline focusing on image quality enhancement using tools.
\ding{184}~\underline{\textbf{Diffusion-based Models}}: We further evaluate how advanced diffusion-based models, \texttt{flux-pro-1.1}~\cite{flux-1d1}, respond to editing instructions and perform corresponding modifications. Note that Google’s NanaBanana and Adobe Firefly are popular image editing solutions but are closed-source. Since their underlying architectures (e.g., pure diffusion or hybrid systems) remain undisclosed, we instead compare against the leading open-source diffusion model, \texttt{flux-pro-1.1}.

\noindent\textbf{Metrics.} Following prior work~\cite{lin2025jarvisart}, we adopt the following metrics:
\ding{182}~L1 and L2 differences between the output and the ground-truth images;
\ding{183}~Perceptual quality and semantic consistency scores evaluated by GLM-4.5V~\cite{hong2025glm}.
For the MIT-Adobe5K dataset, we report PSNR, LPIPS, and SSIM, following the original benchmark setting.

\subsection{Qualitative Comparisons}\label{sec:qualitative}
\begin{figure*}
\centering
\includegraphics[width=0.88\textwidth]{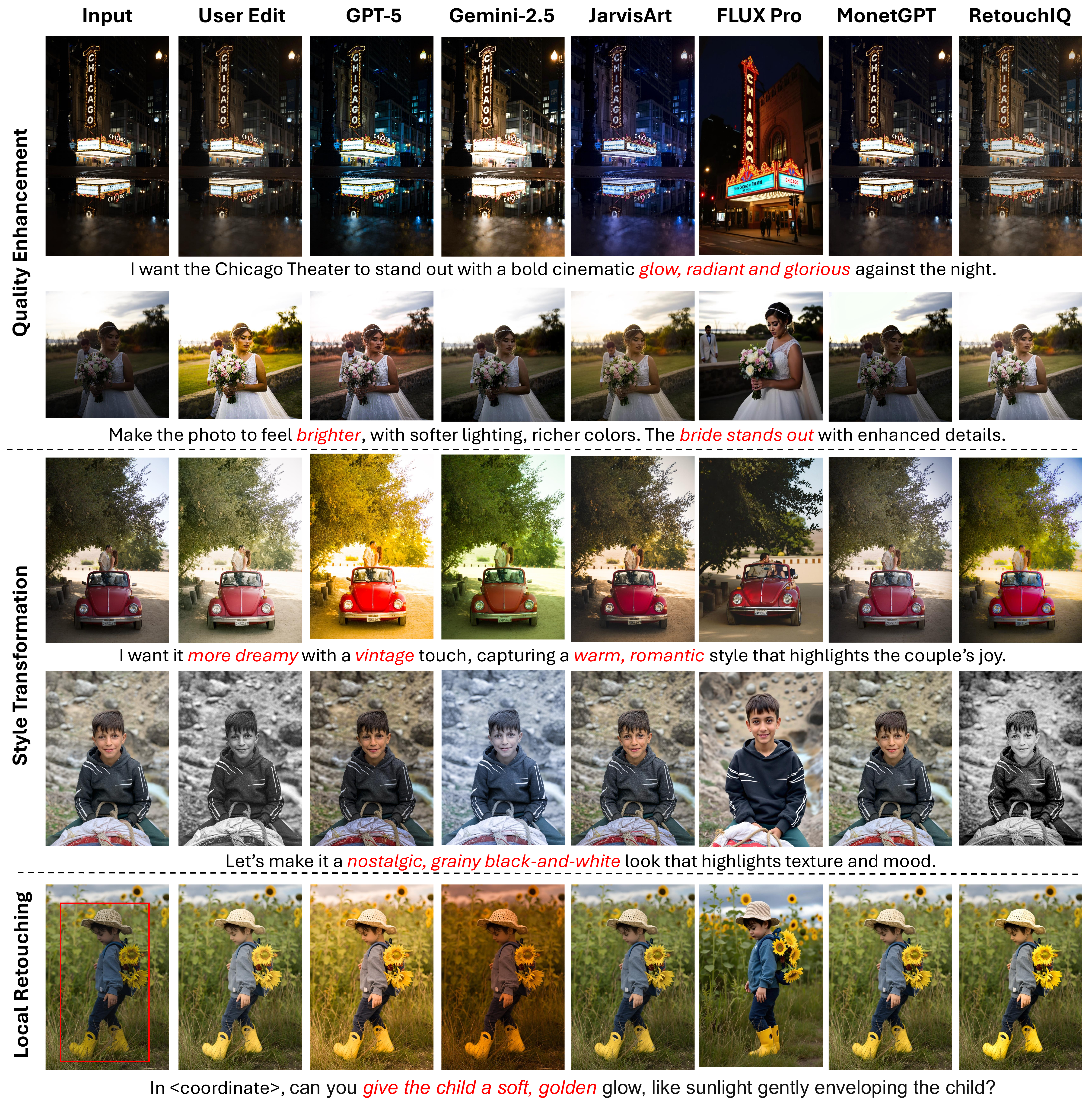}
\vspace*{-0.1in}
\caption{Qualitative results across diverse image retouching scenarios, including quality enhancement (top), style transformation (middle), and local retouching (bottom). For each example, the input image and the corresponding user-edited result are shown on the left.}
\label{fig:exp1}
\vspace*{-0.1in}
\end{figure*}

Figure~\ref{fig:exp1} presents side-by-side comparisons between \textsc{RetouchIQ} and representative baselines across three scenarios: quality improving, style changing, and local retouching.
We observe that each baseline exhibits distinct deficiencies.
First, general-purpose MLLMs such as \texttt{GPT-5} tend to over-edit images, as shown in the 3rd and 5th rows.
This is likely because these models struggle to infer coherent and natural parameter configurations in a zero-shot manner.
Second, the diffusion-based method \texttt{Flux~Pro} fails to preserve the original image structure, leading to noticeable identity and environmental distortions in all examples.
Among MLLM agents, \texttt{MonetGPT} struggles with style changing tasks, while \texttt{JarvisArt} fails on customized requests such as incorrectly adjusts temperatures in 1st row and miss black-and-white conversions in 4th row.
In contrast, \textsc{RetouchIQ} consistently produces results that are more semantically aligned and professionally styled, particularly in challenging scenarios such as nighttime balancing and tone harmonization.

\begin{figure}[t]
    \centering
    \includegraphics[width=0.95\columnwidth]{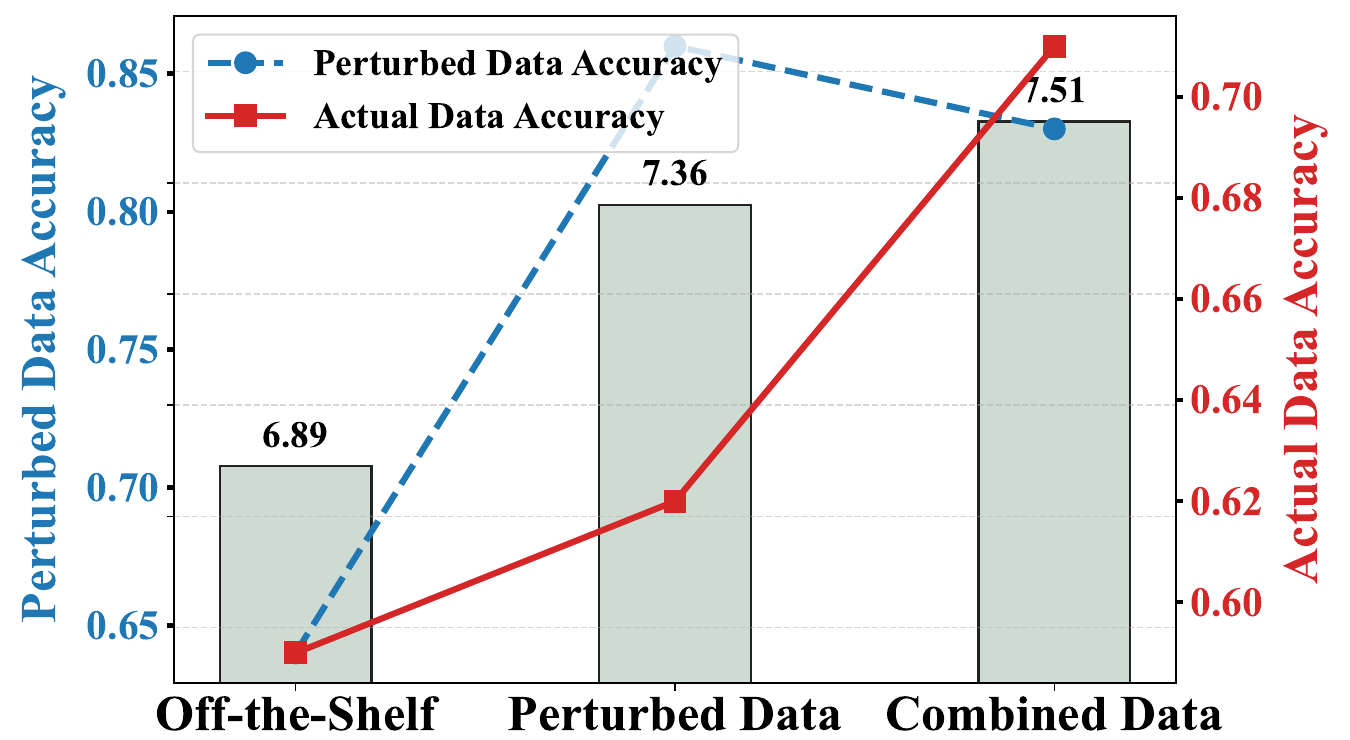}
    \vspace*{-0.1in}
    \caption{Comparison of reward model and policy model performance under different reward model configurations. The lines show the accuracies of the reward model, while the bars indicate the scores of the corresponding policy model.}
    \vspace*{-0.125in}
    \label{fig:reward_model_analysis}
\end{figure}

\begin{table}
    \centering
    \resizebox{0.76\linewidth}{!}{%

\begin{tabular}{lccc}
\toprule

& SSIM & LPIPS & PSNR  \\

\midrule
\textsc{GPT-5}~\cite{gpt5} & 0.72 & 0.26 & 20.82 \\
\textsc{MonetGPT}~\cite{dutt2025monetgpt} & 0.82 & 0.17 & 23.10 \\
\textsc{JarvisArt}~\cite{lin2025jarvisart} & 0.76 & 0.23 & 21.03 \\
\rowcolor{gray!20}
\textsc{RetouchIQ-SFT} & 0.84 & 0.20 & 22.37 \\
\rowcolor{gray!20}
\textsc{RetouchIQ-GRM} & \textbf{0.86} & \textbf{0.16} & \textbf{23.14} \\
\bottomrule

\end{tabular}
}
\caption{Quantitative results on the MIT-Adobe5K benchmark.}
\vspace*{-0.1in}
\label{tab:adobe5k}
\end{table}

\subsection{Quantitative Evaluation}
Next, we quantitatively evaluate performance on RetouchEval and MIT-Adobe5K. The result is in Table \ref{tab:objective_eval} and \ref{tab:adobe5k}. We observe that
RetouchIQ achieves superior results across various scenarios, outperforming diffusion-based and MLLM-agent baselines by large margins. 
Similar gains are observed on MIT-Adobe5K, demonstrating that RetouchIQ generalizes beyond the curated instruction dataset and transfers well to standard aesthetic editing benchmarks.
Notably, the inclusion of RL fine-tuning with the generalist reward contributes a consistent improvement across all metrics, confirming the value of multimodal reward alignment.

\subsection{The Role of Reward Model and PGRT}
In this section, we analyze the role of the reward model in \textsc{RetouchIQ}. Specifically, we examine how policy-guided reward training (PGRT) improves the reward model itself, and how this improvement, in turn, benefits the policy model training. We consider two variants of the generalist reward model: \ding{182} an untrained version, \emph{i.e.}, an off-the-shelf MLLM; and \ding{183} a version trained only on perturbed data generated by the image perturber introduced in Sec.~\ref{sec:GRM}.

The results are in Figure~\ref{fig:reward_model_analysis}. We first report the accuracy of the reward model (line plot). Here, accuracy measures whether the reward model correctly assigns lower rewards to sub-optimal images either from perturbed synthesis data (blue line) or from actual policy-generated data (red line). As shown in the figure, PGRT shifts the reward model’s distribution toward actual policy-generated data, achieving the highest accuracy on that set. Next, we use these reward models to train policy models and report their performance on the \textsc{RetouchEval} benchmark, illustrated by the bar plot. We observe that PGRT yields the best overall performance, indicating that policy-guided reward training effectively produces a reward model that better supports the policy model’s learning process.

\section{Conclusion}
We introduced \textsc{RetouchIQ}, an instruction-based MLLM agent that performs executable image editing via professional software interfaces. By translating natural-language instructions into parameterized edits, it enables controllable, transparent manipulation beyond black-box diffusion methods. A generalist reward model assesses editing quality, providing human-like feedback for reinforcement learning. This enhances instruction consistency and demonstrates the potential of reasoning-based policies with multimodal rewards for intelligent visual editing.

{
    \small
    \bibliographystyle{ieeenat_fullname}
    \bibliography{main}

\begin{thebibliography}{42}
\providecommand{\natexlab}[1]{#1}
\providecommand{\url}[1]{\texttt{#1}}
\expandafter\ifx\csname urlstyle\endcsname\relax
  \providecommand{\doi}[1]{doi: #1}\else
  \providecommand{\doi}{doi: \begingroup \urlstyle{rm}\Url}\fi

\bibitem[Adobe()]{adobe-lightroom}
Adobe.
\newblock Adobe lightroom.
\newblock https://lightroom.adobe.com.

\bibitem[Bai et~al.(2025)Bai, Chen, Liu, Wang, Ge, Song, Dang, Wang, Wang, Tang, et~al.]{bai2025qwen2}
Shuai Bai, Keqin Chen, Xuejing Liu, Jialin Wang, Wenbin Ge, Sibo Song, Kai Dang, Peng Wang, Shijie Wang, Jun Tang, et~al.
\newblock Qwen2. 5-vl technical report.
\newblock \emph{arXiv preprint arXiv:2502.13923}, 2025.

\bibitem[Bychkovsky et~al.(2011)Bychkovsky, Paris, Chan, and Durand]{fivek}
Vladimir Bychkovsky, Sylvain Paris, Eric Chan, and Fr{\'e}do Durand.
\newblock Learning photographic global tonal adjustment with a database of input / output image pairs.
\newblock In \emph{The Twenty-Fourth IEEE Conference on Computer Vision and Pattern Recognition}, 2011.

\bibitem[Chen et~al.(2024)Chen, Li, Gu, Ren, Chen, Ye, Pei, Zhou, Song, and Zhu]{chen2024restoreagent}
Haoyu Chen, Wenbo Li, Jinjin Gu, Jingjing Ren, Sixiang Chen, Tian Ye, Renjing Pei, Kaiwen Zhou, Fenglong Song, and Lei Zhu.
\newblock Restoreagent: Autonomous image restoration agent via multimodal large language models.
\newblock \emph{Advances in Neural Information Processing Systems}, 37:\penalty0 110643--110666, 2024.

\bibitem[{Claid AI}()]{claid}
{Claid AI}.
\newblock {Claid AI Photo Studio}.
\newblock https://claid.ai/.

\bibitem[Comanici et~al.(2025)Comanici, Bieber, Schaekermann, Pasupat, Sachdeva, Dhillon, Blistein, Ram, Zhang, Rosen, et~al.]{comanici2025gemini}
Gheorghe Comanici, Eric Bieber, Mike Schaekermann, Ice Pasupat, Noveen Sachdeva, Inderjit Dhillon, Marcel Blistein, Ori Ram, Dan Zhang, Evan Rosen, et~al.
\newblock Gemini 2.5: Pushing the frontier with advanced reasoning, multimodality, long context, and next generation agentic capabilities.
\newblock \emph{arXiv preprint arXiv:2507.06261}, 2025.

\bibitem[Conde et~al.(2025)Conde, Lu, and Timofte]{conde2025pixtalk}
Marcos~V Conde, Zihao Lu, and Radu Timofte.
\newblock Pixtalk: Controlling photorealistic image processing and editing with language.
\newblock In \emph{Proceedings of the IEEE/CVF International Conference on Computer Vision}, pages 19269--19279, 2025.

\bibitem[Deepmind()]{gemini-agent}
Google Deepmind.
\newblock Introducing gemini 2.0: our new ai model for the agentic era.
\newblock https://blog.google/technology/google-deepmind/google-gemini-ai-update-december-2024/.

\bibitem[Duan et~al.(2025)Duan, Zhang, Lin, Jin, Wang, Zou, Guo, and Li]{duan2025diffretouch}
Zheng-Peng Duan, Jiawei Zhang, Zheng Lin, Xin Jin, XunDong Wang, Dongqing Zou, Chun-Le Guo, and Chongyi Li.
\newblock Diffretouch: Using diffusion to retouch on the shoulder of experts.
\newblock In \emph{Proceedings of the AAAI Conference on Artificial Intelligence}, pages 2825--2833, 2025.

\bibitem[Dutt et~al.(2025)Dutt, Ceylan, and Mitra]{dutt2025monetgpt}
Niladri~Shekhar Dutt, Duygu Ceylan, and Niloy~J Mitra.
\newblock Monetgpt: Solving puzzles enhances mllms' image retouching skills.
\newblock \emph{ACM Transactions on Graphics (TOG)}, 44\penalty0 (4):\penalty0 1--12, 2025.

\bibitem[Fan et~al.(2025)Fan, Feng, Lyu, Zhou, and Yue]{fan2025sophiavl}
Kaixuan Fan, Kaituo Feng, Haoming Lyu, Dongzhan Zhou, and Xiangyu Yue.
\newblock Sophiavl-r1: Reinforcing mllms reasoning with thinking reward.
\newblock \emph{arXiv preprint arXiv:2505.17018}, 2025.

\bibitem[Guo et~al.(2025)Guo, Yang, Zhang, Song, Zhang, Xu, Zhu, Ma, Wang, Bi, et~al.]{guo2025deepseek}
Daya Guo, Dejian Yang, Haowei Zhang, Junxiao Song, Ruoyu Zhang, Runxin Xu, Qihao Zhu, Shirong Ma, Peiyi Wang, Xiao Bi, et~al.
\newblock Deepseek-r1: Incentivizing reasoning capability in llms via reinforcement learning.
\newblock \emph{arXiv preprint arXiv:2501.12948}, 2025.

\bibitem[He et~al.(2024)He, Li, Wang, Zheng, Cao, Yan, Li, Xie, Zhang, and Zhou]{he2024training}
Xuanhua He, Lang Li, Yingying Wang, Hui Zheng, Ke Cao, Keyu Yan, Rui Li, Chengjun Xie, Jie Zhang, and Man Zhou.
\newblock Training-free large model priors for multiple-in-one image restoration.
\newblock \emph{arXiv preprint arXiv:2407.13181}, 2024.

\bibitem[Hertz et~al.(2022)Hertz, Mokady, Tenenbaum, Aberman, Pritch, and Cohen-Or]{hertz2022prompt}
Amir Hertz, Ron Mokady, Jay Tenenbaum, Kfir Aberman, Yael Pritch, and Daniel Cohen-Or.
\newblock Prompt-to-prompt image editing with cross attention control.
\newblock \emph{arXiv preprint arXiv:2208.01626}, 2022.

\bibitem[Ho et~al.(2020)Ho, Jain, and Abbeel]{ho2020denoising}
Jonathan Ho, Ajay Jain, and Pieter Abbeel.
\newblock Denoising diffusion probabilistic models.
\newblock \emph{Advances in neural information processing systems}, 33:\penalty0 6840--6851, 2020.

\bibitem[Hong et~al.(2025)Hong, Yu, Gu, Wang, Gan, Tang, Cheng, Qi, Ji, Pan, et~al.]{hong2025glm}
Wenyi Hong, Wenmeng Yu, Xiaotao Gu, Guo Wang, Guobing Gan, Haomiao Tang, Jiale Cheng, Ji Qi, Junhui Ji, Lihang Pan, et~al.
\newblock Glm-4.5v and glm-4.1v-thinking: Towards versatile multimodal reasoning with scalable reinforcement learning.
\newblock \emph{arXiv e-prints}, pages arXiv--2507, 2025.

\bibitem[Hu et~al.(2018)Hu, He, Xu, Wang, and Lin]{hu2018exposure}
Yuanming Hu, Hao He, Chenxi Xu, Baoyuan Wang, and Stephen Lin.
\newblock Exposure: A white-box photo post-processing framework.
\newblock \emph{ACM Transactions on Graphics (TOG)}, 37\penalty0 (2):\penalty0 1--17, 2018.

\bibitem[Huang et~al.(2024)Huang, Huang, Ning, Lin, Wang, and Liu]{huang2024genmac}
Kaiyi Huang, Yukun Huang, Xuefei Ning, Zinan Lin, Yu Wang, and Xihui Liu.
\newblock Genmac: compositional text-to-video generation with multi-agent collaboration.
\newblock \emph{arXiv preprint arXiv:2412.04440}, 2024.

\bibitem[Ke et~al.(2022)Ke, Sun, Zhu, Xu, and Lau]{ke2022harmonizer}
Zhanghan Ke, Chunyi Sun, Lei Zhu, Ke Xu, and Rynson~WH Lau.
\newblock Harmonizer: Learning to perform white-box image and video harmonization.
\newblock In \emph{European conference on computer vision}, pages 690--706. Springer, 2022.

\bibitem[Kosugi and Yamasaki(2020)]{kosugi2020unpaired}
Satoshi Kosugi and Toshihiko Yamasaki.
\newblock Unpaired image enhancement featuring reinforcement-learning-controlled image editing software.
\newblock In \emph{Proceedings of the AAAI conference on artificial intelligence}, pages 11296--11303, 2020.

\bibitem[Labs()]{flux-1d1}
Black~Forest Labs.
\newblock Flux 1.1 pro.
\newblock https://bfl.ai/models/flux-pro.

\bibitem[Lin et~al.(2025)Lin, Lin, Lin, Bai, Pan, Li, Chen, Wang, Ding, Li, et~al.]{lin2025jarvisart}
Yunlong Lin, Zixu Lin, Kunjie Lin, Jinbin Bai, Panwang Pan, Chenxin Li, Haoyu Chen, Zhongdao Wang, Xinghao Ding, Wenbo Li, et~al.
\newblock Jarvisart: Liberating human artistic creativity via an intelligent photo retouching agent.
\newblock \emph{arXiv preprint arXiv:2506.17612}, 2025.

\bibitem[Liu et~al.(2025{\natexlab{a}})Liu, Ying, Qian, Li, Zhang, Liu, and Zhang]{liu2025mofrr}
Jiaxin Liu, Qichao Ying, Zhenxing Qian, Sheng Li, Runqi Zhang, Jian Liu, and Xinpeng Zhang.
\newblock Mofrr: Mixture of diffusion models for face retouching restoration.
\newblock In \emph{Proceedings of the IEEE/CVF International Conference on Computer Vision}, pages 12842--12851, 2025{\natexlab{a}}.

\bibitem[Liu et~al.(2025{\natexlab{b}})Liu, Wang, Xu, Ma, Ruan, Li, Liu, and Wu]{liu2025inference}
Zijun Liu, Peiyi Wang, Runxin Xu, Shirong Ma, Chong Ruan, Peng Li, Yang Liu, and Yu Wu.
\newblock Inference-time scaling for generalist reward modeling.
\newblock \emph{arXiv preprint arXiv:2504.02495}, 2025{\natexlab{b}}.

\bibitem[Lu et~al.(2025)Lu, Chai, Guo, Yin, Liu, Wang, Xiao, Ren, Xiong, and Li]{lu2025ui}
Zhengxi Lu, Yuxiang Chai, Yaxuan Guo, Xi Yin, Liang Liu, Hao Wang, Han Xiao, Shuai Ren, Guanjing Xiong, and Hongsheng Li.
\newblock Ui-r1: Enhancing efficient action prediction of gui agents by reinforcement learning.
\newblock \emph{arXiv preprint arXiv:2503.21620}, 2025.

\bibitem[Luo et~al.(2025)Luo, Wang, He, Chen, Li, and Xia]{luo2025gui}
Run Luo, Lu Wang, Wanwei He, Longze Chen, Jiaming Li, and Xiaobo Xia.
\newblock Gui-r1: A generalist r1-style vision-language action model for gui agents.
\newblock \emph{arXiv preprint arXiv:2504.10458}, 2025.

\bibitem[OpenAI({\natexlab{a}})]{gpt5}
OpenAI.
\newblock Introducing gpt-5.
\newblock https://openai.com/index/introducing-gpt-5/, {\natexlab{a}}.

\bibitem[OpenAI({\natexlab{b}})]{openai-agent}
OpenAI.
\newblock Introducing chatgpt agent: bridging research and action.
\newblock https://openai.com/index/introducing-chatgpt-agent/, {\natexlab{b}}.

\bibitem[OpenAI({\natexlab{c}})]{openai-deepresearch}
OpenAI.
\newblock Introducing deep research.
\newblock https://openai.com/index/introducing-deep-research/, {\natexlab{c}}.

\bibitem[Ouyang et~al.(2023)Ouyang, Dong, Kang, Ren, Xu, and Xie]{ouyang2023rsfnet}
Wenqi Ouyang, Yi Dong, Xiaoyang Kang, Peiran Ren, Xin Xu, and Xuansong Xie.
\newblock Rsfnet: A white-box image retouching approach using region-specific color filters.
\newblock In \emph{Proceedings of the IEEE/CVF International Conference on Computer Vision}, pages 12160--12169, 2023.

\bibitem[Picsart()]{picsart}
Picsart.
\newblock Picsart.
\newblock https://picsart.com.

\bibitem[Qiao et~al.(2025)Qiao, Tan, Yang, Wang, Wang, Wan, Zhou, Dong, Zeng, Xu, et~al.]{qiao2025we}
Runqi Qiao, Qiuna Tan, Peiqing Yang, Yanzi Wang, Xiaowan Wang, Enhui Wan, Sitong Zhou, Guanting Dong, Yuchen Zeng, Yida Xu, et~al.
\newblock We-math 2.0: A versatile mathbook system for incentivizing visual mathematical reasoning.
\newblock \emph{arXiv preprint arXiv:2508.10433}, 2025.

\bibitem[Shen et~al.(2025)Shen, Pei, Peng, Song, Liu, Peng, Sun, Hao, Wang, Zhang, et~al.]{shen2025skywork}
Wei Shen, Jiangbo Pei, Yi Peng, Xuchen Song, Yang Liu, Jian Peng, Haofeng Sun, Yunzhuo Hao, Peiyu Wang, Jianhao Zhang, et~al.
\newblock Skywork-r1v3 technical report.
\newblock \emph{arXiv preprint arXiv:2507.06167}, 2025.

\bibitem[Shi et~al.(2021)Shi, Xu, Xu, Bui, Dernoncourt, and Xu]{shi2021learning}
Jing Shi, Ning Xu, Yihang Xu, Trung Bui, Franck Dernoncourt, and Chenliang Xu.
\newblock Learning by planning: Language-guided global image editing.
\newblock In \emph{Proceedings of the IEEE/CVF Conference on Computer Vision and Pattern Recognition}, pages 13590--13599, 2021.

\bibitem[Song et~al.(2020)Song, Sohl-Dickstein, Kingma, Kumar, Ermon, and Poole]{song2020score}
Yang Song, Jascha Sohl-Dickstein, Diederik~P Kingma, Abhishek Kumar, Stefano Ermon, and Ben Poole.
\newblock Score-based generative modeling through stochastic differential equations.
\newblock \emph{arXiv preprint arXiv:2011.13456}, 2020.

\bibitem[Wang et~al.(2025{\natexlab{a}})Wang, Qu, Huang, Chu, Lin, and Chen]{wang2025vl}
Haozhe Wang, Chao Qu, Zuming Huang, Wei Chu, Fangzhen Lin, and Wenhu Chen.
\newblock Vl-rethinker: Incentivizing self-reflection of vision-language models with reinforcement learning.
\newblock \emph{arXiv preprint arXiv:2504.08837}, 2025{\natexlab{a}}.

\bibitem[Wang et~al.(2025{\natexlab{b}})Wang, Ding, Zeng, Chen, Chen, Wang, Xie, Huang, and Zhao]{wang2025vrag}
Qiuchen Wang, Ruixue Ding, Yu Zeng, Zehui Chen, Lin Chen, Shihang Wang, Pengjun Xie, Fei Huang, and Feng Zhao.
\newblock Vrag-rl: Empower vision-perception-based rag for visually rich information understanding via iterative reasoning with reinforcement learning.
\newblock \emph{arXiv preprint arXiv:2505.22019}, 2025{\natexlab{b}}.

\bibitem[Wang et~al.(2024)Wang, Li, Li, and Liu]{wang2024genartist}
Zhenyu Wang, Aoxue Li, Zhenguo Li, and Xihui Liu.
\newblock Genartist: Multimodal llm as an agent for unified image generation and editing.
\newblock \emph{Advances in Neural Information Processing Systems}, 37:\penalty0 128374--128395, 2024.

\bibitem[Wu et~al.(2023)Wu, Liu, Zhao, Kale, Bui, Yu, Lin, Zhang, and Chang]{wu2023uncovering}
Qiucheng Wu, Yujian Liu, Handong Zhao, Ajinkya Kale, Trung Bui, Tong Yu, Zhe Lin, Yang Zhang, and Shiyu Chang.
\newblock Uncovering the disentanglement capability in text-to-image diffusion models.
\newblock In \emph{Proceedings of the IEEE/CVF conference on computer vision and pattern recognition}, pages 1900--1910, 2023.

\bibitem[Yang et~al.(2024)Yang, Yu, Meng, Xu, Ermon, and Cui]{yang2024mastering}
Ling Yang, Zhaochen Yu, Chenlin Meng, Minkai Xu, Stefano Ermon, and Bin Cui.
\newblock Mastering text-to-image diffusion: Recaptioning, planning, and generating with multimodal llms.
\newblock In \emph{Forty-first International Conference on Machine Learning}, 2024.

\bibitem[Ye-Bin et~al.(2025)Ye-Bin, Miles, Oh, Elezi, and Deng]{ye2025retouchllm}
Moon Ye-Bin, Roy Miles, Tae-Hyun Oh, Ismail Elezi, and Jiankang Deng.
\newblock Retouchllm: Training-free code-based image retouching with vision language models.
\newblock \emph{arXiv e-prints}, pages arXiv--2510, 2025.

\bibitem[Zhou et~al.(2025)Zhou, Ji, Chen, Sun, Chen, Hong, Han, Guo, and Yang]{zhou2025generative}
Jiayi Zhou, Jiaming Ji, Boyuan Chen, Jiapeng Sun, Wenqi Chen, Donghai Hong, Sirui Han, Yike Guo, and Yaodong Yang.
\newblock Generative rlhf-v: Learning principles from multi-modal human preference.
\newblock \emph{arXiv preprint arXiv:2505.18531}, 2025.

\end{thebibliography}
}

\end{document}